\documentclass[letterpaper, 10 pt, conference]{ieeeconf}  

\IEEEoverridecommandlockouts                              

\usepackage{cite}
\usepackage{amsmath,amssymb,amsfonts}
\usepackage{algorithmic}
\usepackage{graphicx}
\usepackage{textcomp}
\usepackage{xcolor}
\usepackage{tabularx}
\usepackage{multirow}
\usepackage{array}
\usepackage{booktabs}
\def\BibTeX{{\rm B\kern-.05em{\sc i\kern-.025em b}\kern-.08em
    T\kern-.1667em\lower.7ex\hbox{E}\kern-.125emX}}
\begin{document}

\title{ArthroCut: Autonomous Policy Learning for Robotic Bone Resection \\ in Knee Arthroplasty }
\author{Xu Lu$^{\dagger}$, Yiling Zhang$^{\dagger}$, Wenquan Cheng, Longfei Ma, Fang Chen and Hongen Liao* 
\thanks{This work is supported in part by the National Natural Science Foundation of China (U22A2051, 82027807), National Key Research and Development Program of China (2022YFC2405200), the Natural Science Foundation of Beijing Municipality (L252126, L252201), and the Science and Technology Commission of Shanghai Municipality (24511104100).}
\thanks{Xu Lu, Yiling Zhang, Wenquan Cheng, Longfei Ma and Hongen Liao are with the School of Biomedical Engineering, Tsinghua University, Beijing 100084, China.}%
\thanks{Fang Chen and Hongen Liao are with the School of Biomedical Engineering, Shanghai Jiao Tong University, Shanghai 200240, China}
\thanks{Yiling Zhang is also with the Longwood Valley MedTech, Beijing 100176, China.}
\thanks{$^{\dagger}$ Denotes equal contribution}
\thanks{Corresponding author: Hongen Liao, liao@tsinghua.edu.cn.}
}

\maketitle

\begin{abstract}

Despite rapid commercialization of surgical robots, their autonomy and real-time decision-making remain limited in practice. To address this gap, we propose \textit{ArthroCut}, an autonomous policy learning framework that upgrades knee arthroplasty robots from assistive execution to context-aware action generation. ArthroCut fine-tunes a Qwen--VL backbone on a self-built, time-synchronized multimodal dataset from 21 complete cases (23{,}205 RGB--D pairs), integrating preoperative CT/MR, intraoperative NDI tracking of bones and end effector, RGB--D surgical video, robot state, and textual intent. The method operates on two complementary token families---Preoperative Imaging Tokens (PIT) to encode patient-specific anatomy and planned resection planes, and Time-Aligned Surgical Tokens (TAST) to fuse real-time visual, geometric, and kinematic evidence---and emits an interpretable action grammar under grammar/safety-constrained decoding. In bench-top experiments on a knee prosthesis across seven trials, ArthroCut achieves an average success rate of 86\% over the six standard resections, significantly outperforming strong baselines trained under the same protocol. Ablations show that TAST is the principal driver of reliability while PIT provides essential anatomical grounding, and their combination yields the most stable multi-plane execution. These results indicate that aligning preoperative geometry with time-aligned intraoperative perception and translating that alignment into tokenized, constrained actions is an effective path toward robust, interpretable autonomy in orthopedic robotic surgery.
\end{abstract}


\begin{figure}[ht]
\includegraphics[width=8.5cm]{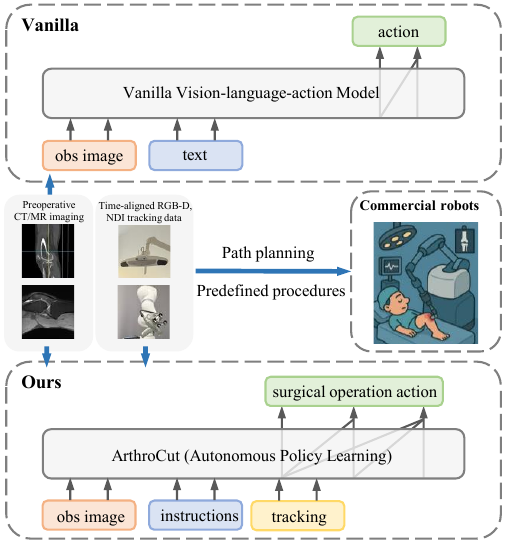}
\caption{\textbf{Comparison between vanilla VLA, commercial orthopedic robots, and the proposed ArthroCut framework.} Conventional VLA models (top) directly map task inputs to robotic actions but fail to incorporate detailed surgical scene understanding, limiting their robustness in complex environments. Commercial orthopedic robots (middle) use intraoperative data to follow pre-templated paths but lack real-time adaptation and personalization~\cite{wang2022progress}. In contrast, ArthroCut (bottom) integrates preoperative imaging with temporally aligned multimodal intraoperative data, enabling accurate, real-time decision-making that improves both surgical precision and personalization. \label{intro}}
\vspace{-0.7em} 
\end{figure}

\section{Introduction}
Surgical robots have transformed many clinical domains by improving precision, reducing variability, and enabling minimally invasive procedures. In orthopedics, robotic systems have become particularly prominent in joint replacement, where millimeter-scale accuracy is critical for implant positioning and long-term outcomes~\cite{zhang2022robotic, zhang2019robotic}. Robotic platforms are now widely adopted in total and partial knee arthroplasty, demonstrating that computer-assisted execution can reduce outliers and improve implant alignment compared to manual techniques~\cite{petrillo2025accuracy}.

Despite these advances, current knee surgery robots are predominantly assistive rather than autonomous~\cite{bautista2019robotics}. They constrain saw paths and guide resections but primarily execute preplanned trajectories under continuous surgeon supervision, with limited ability to perceive intraoperative changes or adapt plans online~\cite{zheng2015computer}. At the same time, the operating room provides rich, complementary signals---including preoperative imaging, optical tracking, and intraoperative surgical video---that are rarely modeled jointly. Harnessing these heterogeneous streams to interpret the scene and generate cutting actions could reduce manual burden, improve consistency, and enable safer, more efficient procedures.

Recent progress in vision--language--action (VLA) learning suggests a path toward such autonomy. Robotics-oriented foundation models now represent actions as tokens, enabling end-to-end policies that map multimodal observations directly to control commands~\cite{zitkovich2023rt, driess2023palm, jiang2022vima}. In parallel, surgical autonomy has progressed from task-level automation to systems capable of language-conditioned, hierarchical policies and recovery from perturbations, with landmark demonstrations of supervised and laparoscopic autonomous soft-tissue anastomosis~\cite{shademan2016supervised, saeidi2022autonomous}. Nevertheless, no prior work has integrated preoperative volumetric imaging, intraoperative surgical scenes, optical tracking, and robot state into a single action-generative model for orthopedic surgery. As shown in Fig.~\ref{intro}, vanilla VLA models (top) adopt a direct mapping strategy from task inputs to robotic actions. While efficient in simplified settings, they lack the ability to parse complex surgical scenes, which severely constrains their robustness and generalizability in high-variability operative environments. Commercial orthopedic robots (middle) represent a step forward by incorporating intraoperative scene information; however, they largely rely on pre-templated surgical paths and standardized execution workflows. As a result, these systems can deliver high geometric accuracy but remain limited in their capacity for real-time adaptation, contextual awareness, and patient-specific customization.
In contrast, the proposed ArthroCut framework (bottom) integrates preoperative imaging with temporally aligned multimodal intraoperative data. This fusion enables the model to continuously contextualize surgical actions within the evolving operative field

To address this gap, we introduce \textbf{ArthroCut}, a multimodal autonomous policy learning framework for robotic bone resection in knee arthroplasty. ArthroCut is trained on a custom dataset that synchronizes preoperative CT/MR volumes, intraoperative NDI pose streams of both patient anatomy and robotic end-effector, RGB-D surgical video, natural language instructions, robot state, and ground-truth motion commands. We fine-tune a GPT-style vision--language model to directly generate robot action primitives. 

Our key contributions conclude:
\begin{itemize}
  \item We construct the first aligned, multimodal dataset for autonomous knee arthroplasty, pairing \textbf{21} patients’ preoperative CT/MR volumes with \textbf{23{,}205} RGB–D image pairs and \textbf{21} full-length intraoperative NDI tracking logs, alongside robot state, language intent, and action commands.
  \item We develop a VLA–based policy that fuses Preoperative Imaging Tokens (PIT) and Time-Aligned Surgical Tokens (TAST) and autoregressively generates textual motion instructions via an interpretable action grammar with quantized parameters under grammar/safety-constrained decoding.
  \item In benchtop tests on a knee prosthesis (six resections), we attains a SR of \textbf{0.86} and SPL of \textbf{0.75} across the six resections, and reduce robot execution time by \textbf{11.95\%} vs. a vanilla baseline under identical constraints, improving accuracy and throughput for sequential, multi-plane osteotomies.
\end{itemize}

\section{Related work}\label{problem}

\subsection{Orthopedic surgical robots}

Robotic assistance in knee arthroplasty has been widely investigated as a means to enhance surgical precision, reproducibility, and patient-specific outcomes. Numerous clinical studies have demonstrated that robotic systems improve component positioning and reduce alignment outliers relative to conventional jig-based techniques~\cite{kayani2018iatrogenic}. Both image-based and imageless platforms have shown improved accuracy in achieving mechanical alignment and restoring limb axes in total knee arthroplasty (TKA)~\cite{bautista2019robotics}. Similarly, robotic-assisted unicompartmental knee arthroplasty (UKA) has been associated with improved functional outcomes, faster recovery, and radiographic precision comparable or superior to manual UKA~\cite{kayani2019robotic, sun2021does}. Furthermore, comparative analyses suggest that robotic systems outperform navigation-assisted workflows in coronal-plane accuracy, highlighting their potential to reduce intraoperative variability~\cite{figueroa2019comparison}. 

Beyond alignment, robotic systems may also contribute to reduced intraoperative bone resection errors, more consistent soft-tissue balancing, and a lower incidence of revision surgery in selected patient cohorts~\cite{bayoumi2023ten}. Nevertheless, these benefits come with limitations: existing platforms typically rely on constraint-based execution of preplanned cuts, offer limited intraoperative perception, and rarely integrate heterogeneous data streams such as volumetric imaging, real-time tracking, and video. This lack of multimodal integration constrains adaptability when intraoperative conditions deviate from preoperative plans. By contrast, our approach explicitly fuses preoperative CT/MR with intraoperative RGB-D, optical tracking, and robot state to generate verifiable action primitives that can adapt trajectories online, moving toward autonomy that is both data-driven and clinically grounded.

\begin{figure*}[t]
\includegraphics[width=\textwidth]{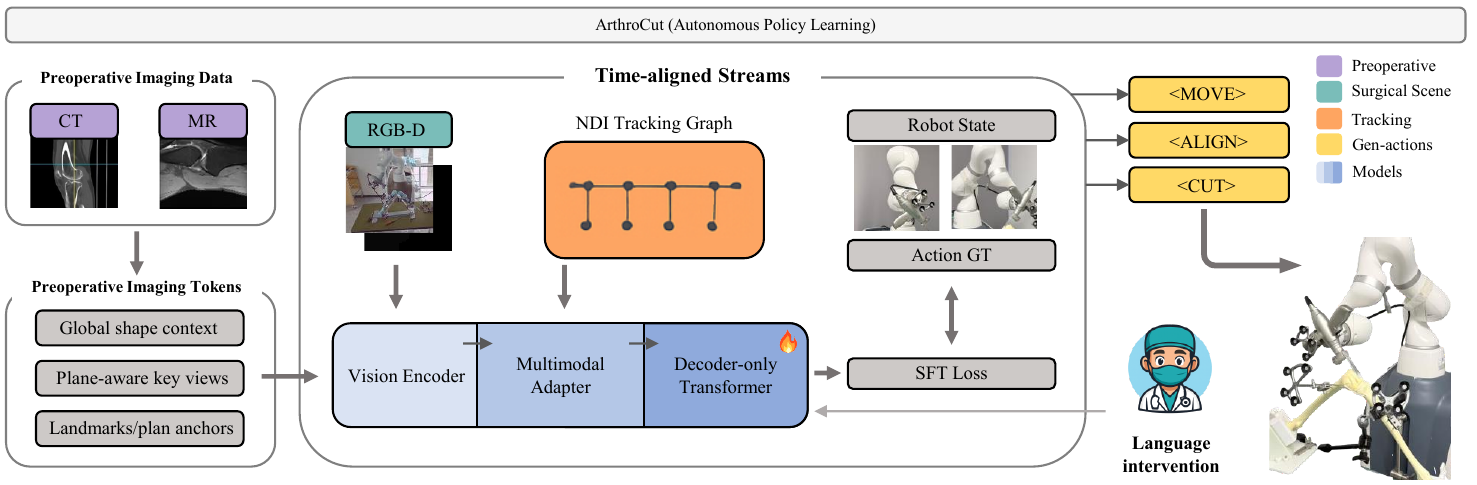}
\caption{\textbf{The overview of our proposed method for robotic bone resection in knee arthroplasty.} Preoperative imaging and intraoperative multimodal streams are encoded into tokens and processed by a transformer-based model to generate surgical actions ($\texttt{<MOVE>}, \texttt{<ALIGN>}, \texttt{<CUT>}$) for autonomous femoral cutting.\label{overview}}
\end{figure*}

\subsection{Vision–-Language–-Action models for robotics}

Vision--language--action (VLA) learning frames robot control as sequence modeling over multimodal tokens, enabling end-to-end policies that directly map complex sensory inputs to action outputs~\cite{kim2024openvla, intelligence2025pi_, zhao2025cot, zhen20243d, guo2025improving, ding2024quar, gbagbe2024bi, wen2025tinyvla, shukor2025smolvla, zhang2025mole, liu2025hybridvla}. Robotics transformers co-fine-tuned on web-scale vision--language datasets and robot trajectories, such as RT-2, demonstrate semantic generalization and the ability to generate action tokens grounded in high-level instructions~\cite{zitkovich2023rt}. Embodied multimodal large language models (e.g., PaLM-E) further show how continuous sensory inputs (RGB, depth, proprioception) can be interleaved with language to support perception, reasoning, and control in a unified framework~\cite{driess2023palm}. Multimodal prompting frameworks (VIMA) extend this paradigm by conditioning robot actions on diverse input modalities, enabling compositional task execution~\cite{jiang2022vima}. Program-synthesis methods (e.g., Code-as-Policies) leverage LLMs to generate executable action code, broadening the expressivity of robot policies~\cite{liang2022code}. In addition, affordance-aware LLM planning approaches couple LLMs with symbolic planners and affordance detectors, thereby improving long-horizon reasoning and task success~\cite{birr2024autogpt+}. 

Despite these advances, most VLA systems remain focused on tabletop manipulation and household robotics benchmarks, with limited consideration for high-stakes medical environments. Critically, they seldom incorporate preoperative volumetric medical imaging or enforce geometry-grounded safety constraints, which are essential in surgical applications. Our framework addresses this gap by combining cross-modal 3D anatomical grounding with tokenized surgical motion primitives, explicitly tailored to multi-plane bone resections in orthopedic surgery.

\subsection{Autonomy in surgical robotics}

The spectrum of autonomy in surgical robotics ranges from pure teleoperation to shared control, conditional autonomy, and fully autonomous execution\cite{attanasio2021autonomy}. Landmark demonstrations include the Smart Tissue Autonomous Robot (STAR), which achieved supervised autonomous soft-tissue suturing in vivo with superior consistency to human surgeons~\cite{shademan2016supervised}, and the demonstration of autonomous laparoscopic small-bowel anastomosis capable of perturbation recovery in porcine models~\cite{saeidi2022autonomous}. These milestones underscore the feasibility of autonomy in surgical contexts, particularly in soft-tissue procedures. Authoritative surveys and perspectives from leading journals emphasize the importance of safety assurance, rigorous verification/validation protocols, and maintaining meaningful human oversight as autonomy levels increase~\cite{yang2017medical}. More recent frameworks such as SRT-H propose hierarchical policies conditioned on language, enabling adaptive stepwise planning and partial autonomy in minimally invasive procedures~\cite{kim2025srt}.

Nevertheless, prior work has largely concentrated on soft-tissue surgery and task-specific pipelines, with limited generalization to orthopedic procedures requiring rigid bone resection. Moreover, most approaches have not attempted to unify preoperative volumetric imaging with intraoperative tracking and multimodal sensing for real-time decision-making. In contrast, our method integrates multimodal information streams into an action-generative policy that produces verifiable motion primitives, enabling online adaptation and enhancing safety in orthopedic surgical autonomy.

\section{Method}


In this section, we introduce \textbf{ArthroCut}, an autonomous policy-learning framework for robotic bone resection in knee arthroplasty. As shown in Fig.~\ref{overview}, ArthroCut converts two input families into text-compatible tokens: (i) \emph{Preoperative Imaging Tokens (PIT)} that summarize CT/MR anatomy and surgical intent (global shape, plane-aware key views, landmark/plan anchors), and (ii) \emph{Time-Aligned Surgical Tokens (TAST)} that fuse synchronized intraoperative streams—RGB-D features, an SE(3) graph of NDI-tracked end-effector/bone poses, and robot kinematic/state vectors. A vision encoder and multimodal adapter interleave PIT and TAST, while a decoder-only transformer autoregressively emits a structured action string via an Action Grammar Tokenizer (e.g., $\texttt{<MOVE>}, \texttt{<ALIGN>}, \texttt{<CUT>}$ with quantized parameters). A constraint decoder applies grammar and safety masks to ensure feasible, verifiable commands for multi-plane resections. Training uses next-token supervision with lightweight auxiliaries (plane grounding, SE(3) consistency, motion smoothness), aligning preoperative geometry with intraoperative evidence and yielding interpretable, stepwise policies.

\subsection{The Base Vision--Language--Action Model}
\label{base model}
We adopt Qwen2.5–VL–32B–Instruct~\cite{bai2025qwen2} as the foundational multimodal policy backbone. Conceptually, the model follows a modern vision–language (VL) architecture: a high-capacity visual encoder $\phi_{\text{vis}}(\cdot)$ processes images or video frames into patch-level features; a lightweight multimodal adaptor $\psi(\cdot)$ projects these features into the shared language-embedding space; and a decoder-only transformer $\mathcal{L}_{\theta}$ (a 32B-parameter autoregressive language backbone) performs unified reasoning over interleaved visual tokens and text tokens. The model has been instruction-tuned to follow natural-language prompts while grounding responses in visual evidence, supporting both multi-image inputs and temporally ordered frame batches. Inference is carried out via standard next-token prediction under a causal objective.

We select Qwen2.5–VL–32B for three reasons. First, it combines strong instruction following and robust multimodal grounding in a single, self-hostable framework. This makes it well suited to surgical settings that require long, interleaved inputs—preoperative CT/MR, intraoperative RGB-D, tracker states, and live robot telemetry. Second, its vision–language pretraining provides transferable capabilities in OCR, spatial reasoning, and layout interpretation that are advantageous in operating-room environments (e.g., labels, rulers, guides, implants). Instruction tuning further aligns naturally with our textual action grammar, enabling stepwise, interpretable motion commands rather than opaque latent controls. Third, the 32B-parameter scale provides sufficient capacity for multi-plane geometric reasoning and temporal consistency without brittle task-specific heuristics. In practice, its mature tokenizer/IO stack, multi-image support, and stable generation controls (temperature, stop tokens) facilitate the seamless incorporation of surgical constraints (e.g., alignment checks, margin tolerances, safety pauses) directly into prompt templates.

\subsection{Token Representation for Time-Aligned Multimodal Surgical Data}
\label{tokens}

We represent all inputs to the VLA backbone as \emph{text-compatible tokens} that are compact, interpretable, and capable of preserving geometry, appearance, and control semantics. By synchronizing all modalities on a common timeline, the model can jointly reason over both preoperative and intraoperative evidence.  

\subsubsection{Preoperative Imaging Tokens (PIT)}

Preoperative CT/MR volumes $V^{\text{CT}}, V^{\text{MR}}$ are summarized into three complementary token sets.  

(1) \textbf{Global shape context:}  
A 3D encoder $\phi_{\text{3D}}$ produces a pooled descriptor capturing overall anatomy:  
\begin{equation}
\mathbf{c}_{\text{glob}} = \operatorname{pool}\!\big(\phi_{\text{3D}}(V^{\text{CT}}, V^{\text{MR}})\big).
\end{equation}

(2) \textbf{Plane-aware views:}  
For each planned resection plane $P_m$, a small set of aligned slices $\{I^{(m)}_k\}$ are encoded and pooled:  
\begin{equation}
\mathbf{c}^{(m)}_{\text{view}} = \operatorname{pool}\!\big(\{\phi_{\text{vis}}(I^{(m)}_k)\}\big).
\end{equation}

(3) \textbf{Landmark/plan anchors:}  
Key anatomical landmarks $\ell_i$ and plane parameters $(\mathbf{n}_m, b_m)$ are discretized into symbolic bins:  
\begin{equation}
\tilde{\ell}_i = \mathrm{bin}(\ell_i), \quad \tilde{\mathbf{n}}_m = \mathrm{bin}(\mathbf{n}_m), \quad \tilde{b}_m = \mathrm{bin}(b_m).
\end{equation}

All tokens are serialized into a compact block $\mathbf{C}$ that encodes both global anatomy and surgical intent.  

\subsubsection{Time-Aligned Surgical Tokens (TAST)}

Intraoperative data streams—RGB-D frames, optical tracking states, and robot kinematics—are aligned on a shared timeline $\mathcal{T}$. Each step $t$ collects a local observation window $W_t$ to capture short-term dynamics.  

(1) \textbf{RGB-D scene encoder:}  
Stacked multi-frame inputs yield visual tokens:  
\begin{equation}
\mathbf{v}_t = \phi_{\text{vis}}(\{I_\tau \oplus D_\tau\}_{\tau \in W_t}).
\end{equation}

(2) \textbf{SE(3) graph encoder:}  
Relative transforms between tracked objects (end-effector, femur, tibia) are embedded as graph features:  
\begin{equation}
\mathbf{n}_t = \mathrm{S3GE}(\{\,T^{o\to o'}_t\,\}).
\end{equation}

(3) \textbf{Robot state encoder:}  
Joint angles $\mathbf{q}_t$, velocities $\dot{\mathbf{q}}_t$, and torques $\boldsymbol{\tau}_t$ are mapped into state tokens:  
\begin{equation}
\mathbf{r}_t = \mathrm{ASE}(\mathbf{q}_t, \dot{\mathbf{q}}_t, \boldsymbol{\tau}_t).
\end{equation}

(4) \textbf{Action grammar tokenizer:}  
Ground-truth commands are serialized into motion primitives:  
\begin{equation}
\mathbf{y}_t = \texttt{<MOVE>}[\tilde{\mathbf{p}}]~\texttt{<ALIGN>}[\tilde{\pi}]~\texttt{<CUT>}[\tilde{s}].
\end{equation}

Finally, the model input at time $t$ concatenates preoperative context with intraoperative observations and the previous action:  
\begin{equation}
\mathbf{Z}_t = \operatorname{concat}(\mathbf{C},~\mathbf{v}_t,~\mathbf{n}_t,~\mathbf{r}_t,~\mathbf{y}_{t-1}).
\end{equation}

This formulation fuses patient-specific anatomy with real-time surgical evidence, enabling autoregressive generation of structured, interpretable action sequences.

\begin{figure*}[t]
\includegraphics[width=\textwidth]{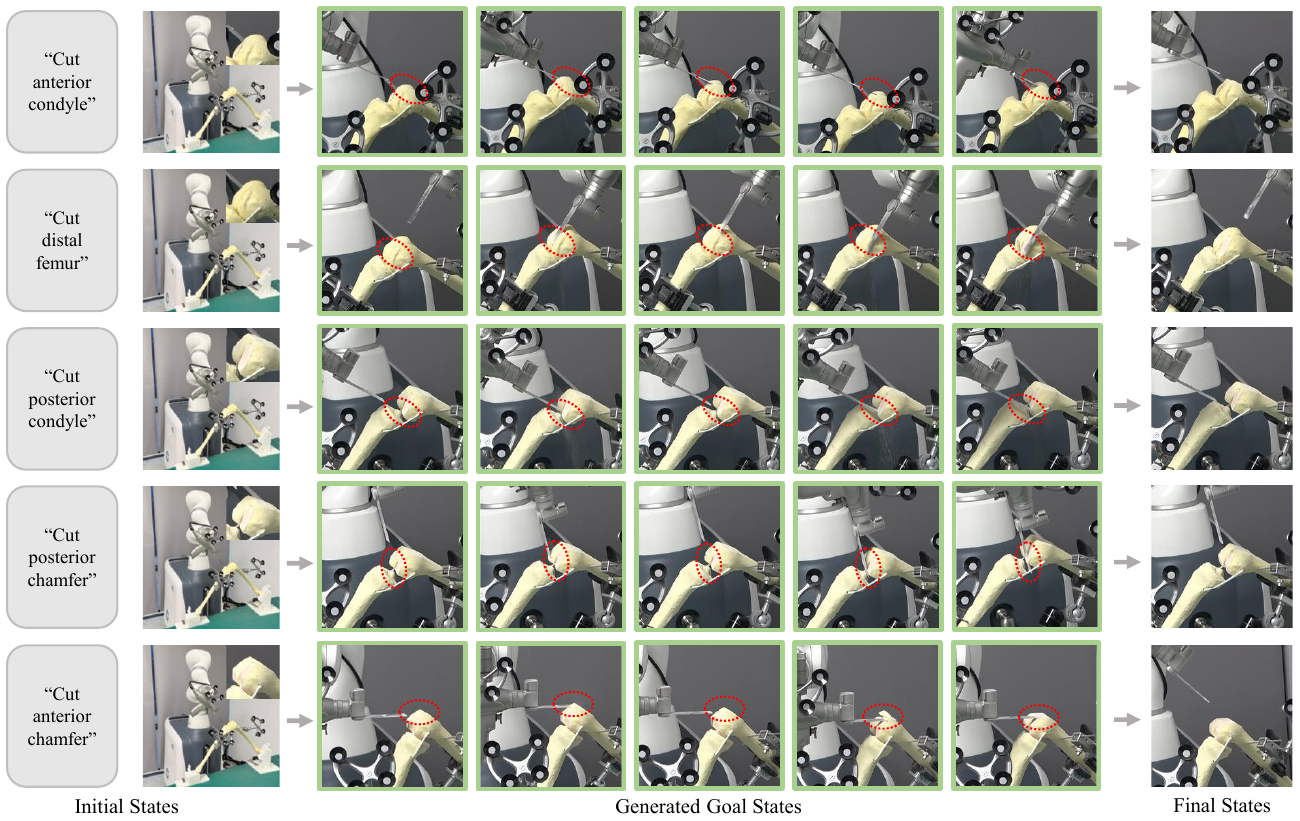}
\caption{\textbf{Example of task execution for femoral cutting across five planes using ArthroCut.} For each task: Left—text instruction with the initial state (top-right inset of the second column shows the pre-cut bone surface); Middle—generated intermediate goal states; Right—final state upon task completion. Full execution trajectories are provided in the supplementary video.\label{results}}
\end{figure*}

\subsection{Training Procedures}
\label{training}


\textbf{Setup and objective.}
We fine-tune Qwen2.5--VL--32B--Instruct to autoregressively emit action tokens $y_{t:t+H}$ given a serialized prefix that concatenates the preoperative context $\mathbf{C}$, the current observation block $\mathbf{X}_t$, and preceding actions (cf.~\ref{tokens}). Actions are represented by an Action Grammar Tokenizer (AGT) that encodes motion primitives with quantized parameters. Training uses teacher forcing, with supervision applied only on action tokens; observation tokens act as fixed prefixes. The primary loss is next-token SFT:
\begin{equation}
\mathcal{L}_{\text{SFT}}
= - \!\!\sum_{\tau=t}^{t+H}\! \log p_{\theta}\!\big(y_{\tau}\mid \text{prefix},\, y_{t:\tau-1}\big),
\end{equation}
optionally combined with auxiliary terms (e.g., plane grounding, SE(3) consistency, action smoothness):
\[
\mathcal{L}_{\text{total}}=\mathcal{L}_{\text{SFT}}+\sum_i \lambda_i \mathcal{L}_i.
\]
Decoding is unchanged by auxiliaries.



\textbf{Sequence packing and prompts.}
Each instance follows a compact instruction template that declares modality tags and task intent, then interleaves $\mathbf{C}$, $\mathbf{X}_t$, and the previous action block. To maximize throughput, multiple temporal windows are concatenated up to the context budget with \texttt{<SEP>} delimiters to avoid cross-window leakage. All numeric fields use the same quantization bins as at inference to prevent train–test mismatch. When latency compensation is needed, targets are shifted by $\Delta t$ to align with system delay.

\textbf{Safety-aware token masking.}
We apply grammar-constrained masking during training and decoding to enforce semantic and physical validity. The mask suppresses infeasible primitives (e.g., disallow \texttt{<CUT>} before a valid \texttt{<ALIGN>}) and invalid parameter bins conditioned on instantaneous state and plan, while preserving gradients over admissible choices. Using the same constraints at inference ensures train–deployment consistency and restricts generation to the feasible action space.

\section{Experiments}

\subsection{Dataset Creation}

We collected a patient-level, time-synchronized multimodal dataset from
\textbf{21} complete primary knee arthroplasty procedures performed with a commercial orthopedic robot
under attending surgeons. Each case includes a full intraoperative timeline and paired preoperative
CT--MR imaging for the operated knee. During surgery, six clinically standard resections are performed:
tibial, distal femur, anterior condyle, posterior condyle,
anterior chamfer, and posterior chamfer.
These phases provide natural supervision for multi-plane planning and execution.

An RGB--D camera records the surgical field, yielding \textbf{23{,}205} synchronized RGB/Depth pairs
across all cases.
Intraoperative optical tracking (NDI) logs the SE(3) poses of the end-effector, femur, and tibia;
the robot controller logs joint state, velocities/torques and tool–center-point pose;
and the console exports time-stamped surgical events and motion commands.
All streams are hardware time-stamped and \emph{software-aligned} onto a common reference grid
($25$\,ms step) before tokenization (cf. \ref{tokens}).

This dataset is designed to (i) ground preoperative geometry and plan intent (PIT) and
(ii) train a time-aligned, multimodal policy to generate tokenized robot actions for
multi-plane bone resections. It provides the necessary coverage of tibial and femoral cuts as well as
anterior/posterior condyle and chamfer planes to evaluate \emph{sequential} planning and execution.

\begin{table*}[t]
\centering
\caption{\textbf{Bench evaluation on a knee prosthesis (SR).} Success threshold $\delta{=}1.5$\,mm mean surface deviation across the six planes. All methods use the same action grammar and constrained decoding. ft: fine-tune.}
\label{tab:main_results}
\renewcommand{\arraystretch}{1.3}
\resizebox{\textwidth}{!}{%
\begin{tabular}{l|cccccc}
\hline
\multicolumn{1}{c|}{\multirow{3}{*}{Method}} & \multicolumn{6}{c}{Metrics}                                                                             \\ \cline{2-7} 
\multicolumn{1}{c|}{}                        & \multicolumn{6}{c}{Success Rate (SR) $\uparrow$}                                                                       \\ \cline{2-7} 
\multicolumn{1}{c|}{}                        &  anterior chamfer    & distal femur & anterior condyle & posterior condyle & tibial & posterior chamfer \\ \hline
OpenVLA (ft)~\cite{kim2024openvla}                                 & 0.29 (2/7) & 0.29 (2/7)   & 0.29 (2/7)       & 0.00 (0/7)        & 0.29 (2/7)       & 0.00 (0/7)        \\
Diffusion Policy~\cite{chi2023diffusion}                             & 0.47 (4/7) & 0.29 (2/7)   & 0.47 (4/7)       & 0.29 (2/7)        & 0.00 (0/7)       & 0.00 (0/7)        \\
RT-2 (ft)~\cite{zitkovich2023rt}                                    & 0.29 (2/7) & 0.29 (2/7)   & 0.29 (2/7)       & 0.00 (0/7)        & 0.00 (0/7)       & 0.00 (0/7)        \\
Octo (ft)~\cite{team2024octo}                                    & 0.29 (2/7) & 0.00 (0/7)   & 0.00 (0/7)       & 0.29 (2/7)        & 0.00 (0/7)       & 0.00 (0/7)        \\
Qwen2.5-VL-Instruct (vanilla)~\cite{bai2025qwen2}                & 0.00 (0/7) & 0.00 (0/7)   & 0.29 (2/7)       & 0.00 (0/7)        & 0.00 (0/7)       & 0.00 (0/7)        \\
ArthroCut (Ours)                             & \textbf{1.00} (7/7) & \textbf{1.00} (7/7)   & \textbf{1.00} (7/7)       & \textbf{1.00} (7/7)        & \textbf{0.57} (4/7)       & \textbf{0.57} (4/7)        \\ \hline
\end{tabular}
}

\end{table*}

\begin{table*}[h]
\centering
\caption{\textbf{Bench evaluation on a knee prosthesis (SPL).} Success threshold $\delta{=}1.5$\,mm mean surface deviation across the six planes. SPL is computed with the protocol and reported as mean $\pm$$\sim$SD over episodes. All methods use the same action grammar and constrained decoding. ft: fine-tune.}
\label{tab:main_results2}
\renewcommand{\arraystretch}{1.3}
\resizebox{\textwidth}{!}{%
\begin{tabular}{l|cccccc}
\hline
\multicolumn{1}{c|}{\multirow{3}{*}{Method}} & \multicolumn{6}{c}{Metrics}                                                                                                                                                                                                                                           \\ \cline{2-7} 
\multicolumn{1}{c|}{}                        & \multicolumn{6}{c}{Success weighted by Path Length (SPL) $\uparrow$}                                                                                                                                                                                                         \\ \cline{2-7} 
\multicolumn{1}{c|}{}                        & anterior chamfer                                    & distal femur                              & anterior condyle                          & posterior condyle                         &  tibial                         & posterior chamfer                         \\ \hline
OpenVLA (ft)~\cite{kim2024openvla}                                 & 0.22 $\pm$ 0.02                           & 0.24 $\pm$ 0.02                           & 0.24 $\pm$ 0.02                           & 0.00 $\pm$ 0.00                           & 0.20 $\pm$ 0.04                           & 0.00 $\pm$ 0.00                           \\
Diffusion Policy~\cite{chi2023diffusion}                             & 0.42 $\pm$ 0.05                           & 0.23 $\pm$ 0.03                           & 0.43 $\pm$ 0.04                           & 0.22 $\pm$ 0.03                           & 0.00 $\pm$ 0.00                           & 0.00 $\pm$ 0.00                           \\
RT-2 (ft)~\cite{zitkovich2023rt}                                    & 0.21 $\pm$ 0.02                           & 0.18 $\pm$ 0.03                           & 0.20 $\pm$ 0.03                           & 0.00 $\pm$ 0.00                           & 0.00 $\pm$ 0.00                           & 0.00 $\pm$ 0.00                           \\
Octo (ft)~\cite{team2024octo}                                    & 0.22 $\pm$ 0.03                           & 0.00 $\pm$ 0.00                           & 0.00 $\pm$ 0.00                           & 0.22 $\pm$ 0.02                           & 0.00 $\pm$ 0.00                           & 0.00 $\pm$ 0.00                           \\ 
Qwen2.5-VL-Instruct (vanilla)~\cite{bai2025qwen2}                & 0.00 $\pm$ 0.00                           & 0.00 $\pm$ 0.00                           & 0.24 $\pm$ 0.03                           & 0.00 $\pm$ 0.00                           & 0.00 $\pm$ 0.00                           & 0.00 $\pm$ 0.00                           \\
ArthroCut (Ours)                             & \textbf{0.92 $\pm$ 0.03} & \textbf{0.93 $\pm$ 0.02} & \textbf{0.95 $\pm$ 0.02} & \textbf{0.91 $\pm$ 0.04} & \textbf{0.39 $\pm$ 0.08} & \textbf{0.42 $\pm$ 0.07} \\ \hline
\end{tabular}
}

\end{table*}

\subsection{Evaluation Metrics}
\label{metrics}

We adopt a single, task-level success criterion and report two metrics: \emph{Success Rate (SR)} and \emph{Success weighted by Path Length (SPL)}. An episode is deemed successful ($S_i{=}1$) if all six resections (tibial, distal femur, anterior/posterior condyle, anterior/posterior chamfer) are executed and the mean surface deviation between the executed and planned cut surfaces, averaged over the six planes, is within a fixed tolerance $\delta$\,mm; otherwise $S_i{=}0$. The surface deviation per plane is computed as the bidirectional Chamfer distance (in mm) between uniformly sampled points on the executed and planned surface patches within the resection window.

\subsection{Evaluations Results}

We evaluate on a \emph{physical bench test} using a knee prosthesis and bone analogs, executing the complete sequence of six resections over \textbf{7} independent runs. Success is defined in \ref{metrics}: an episode is successful if, after all six planes are cut, the \emph{mean surface deviation} from plan (bidirectional Chamfer distance over the resection patches) averaged across the six planes is $\le \delta{=}1.5$\,mm. We report \textbf{SR} and \textbf{SPL} with grammar/safety-constrained decoding for all methods. For fairness, all baselines are retrained on our dataset with the same train/val split and receive the same action grammar; decoding uses identical constraints and temperature/top-$p$ settings.

\noindent\textbf{Baselines:}
We compare against five strong policies: (1) \textbf{Diffusion Policy} (pixel-to-action diffusion with multi-frame RGB-D)~\cite{chi2023diffusion}   , (2) \textbf{OpenVLA} (open-source VLA backbone fine-tuned on our tokens)~\cite{kim2024openvla} , (3) \textbf{Octo} (large multi-task policy with vision-language prompts)~\cite{team2024octo}, (4) \textbf{RT-2} (vision-language-action transformer with text-as-action finetuning)~\cite{zitkovich2023rt}   , and (5) \textbf{Qwen2.5-VL-Instruct (vanilla)}~\cite{bai2025qwen2}—a backbone-matched baseline trained with our grammar but \emph{without} PIT or the structure-aware S3GE encoder or safety masks. Our method \textbf{ArthroCut (Ours)} uses PIT+TAST tokens, S3GE, grammar masks, and the training protocol in \ref{training}.

As shown in Table~\ref{tab:main_results},~\ref{tab:main_results2}, ArthroCut achieves the highest SR and SPL, attaining a 100\% success rate across four primary osteotomy planes—anterior chamfer, distal femur, anterior condyle, and posterior condyle. Performance decreases on the anterior and posterior chamfer planes, where the minimal bone resection leads to prosthesis instability and consequently lower success rates. Among competing methods, the strongest baseline is the Diffusion Policy, while diffusion-style and generalist VLA approaches (e.g., RT-2, Octo) occasionally reach individual planes but frequently drift or require detours, impairing both reliability and efficiency. The vanilla Qwen2.5 baseline rarely completes tasks, highlighting the critical role of our key innovations: PIT tokens (preoperative geometry), structure-aware tracking (S3GE), and grammar/safety masks for converting multimodal evidence into safe and executable cutting actions. Owing to its superior success rate across osteotomy tasks, ArthroCut also achieves the highest SPL among all evaluated methods.

\subsection{Ablation Study}

We introduce two complementary token families to bridge planning and execution in knee arthroplasty. Preoperative Imaging Tokens (PIT) provides a stable anatomical prior that reduces ambiguity in plane identification and aligns action generation with the preplanned resection objectives. Time-Aligned Surgical Tokens (TAST) fuse synchronized intraoperative evidence—RGB-D scene features, an SE(3) graph of NDI-tracked poses for end effector and bones, robot kinematic/state vectors, and the previous action—so the policy can reason over real-time, structure-aware relations and temporal continuity.

As shown in Table~\ref{ab_study}, the introduction of time-aligned surgical tokens (TAST) emerges as the dominant factor driving performance, elevating the vanilla policy from \(\mathrm{SR}=0.05/\mathrm{SPL}=0.04\) to \(0.68/0.61\) (+0.63 and +0.57 absolute). Incorporating preoperative imaging tokens (PIT) alone provides a modest yet consistent improvement (\(0.12/0.10\), +0.07/+0.06), reflecting the utility of anatomical priors even in isolation. When combined (ArthroCut), the two components achieve the highest performance with \(\mathrm{SR}=0.86\) and \(\mathrm{SPL}=0.75\), representing gains of +0.18/+0.14 over TAST alone and +0.81/+0.71 over vanilla. These results highlight a synergistic effect: PIT anchors the model to planned anatomy and target planes, while TAST delivers real-time, structure-aware cues essential for robust and efficient multi-plane cutting.

\subsection{Arthroplasty Verification and Efficiency Analysis}

\begin{figure}[t]
    \centering
    \includegraphics[width=7cm]{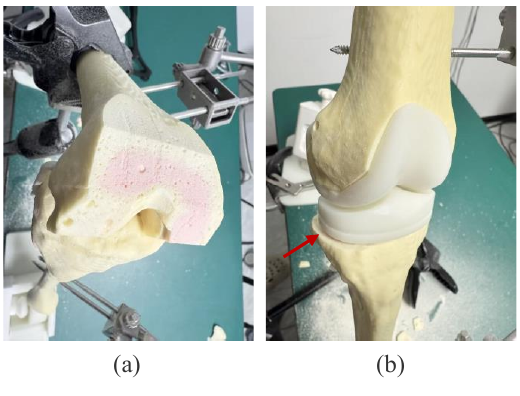}
    \vspace{-0.5em}
    \caption{\textbf{Schematic diagram of knee replacement verification after osteotomy.} (a) Femoral resections provide a visually flat surface for component seating; (b) the joint shown in full extension after component placement. \label{arthro_verif}}
    \vspace{-0.7em} 
\end{figure}

Beyond the quantitative success criteria, we qualitatively verified prosthesis fit on the post-cut knee surrogates. As illustrated in Fig.~\ref{arthro_verif}(a), the femoral resections seat the component with visually flat, coplanar surfaces and negligible step-offs across cut transitions, consistent with the high per-plane completion rates. Figure~\ref{arthro_verif}(b) shows the joint in full extension after component placement—a configuration critical for stability during weight bearing and gait. In this posture, the femoral component sits flush, whereas a small gap is visible at the tibial tray–bone interface. This observation accords with our per-plane statistics, where the tibial plane exhibits the lowest success rate among the six tasks.

Table~\ref{tab:efficiency} summarizes policy latency and robot execution time. With grammar- and safety-constrained decoding, ArthroCut achieves a mean per-action latency of $302.52$ ms on a single NVIDIA~A100; the dominant contribution to wall-clock time remains mechanical motion and material removal. Because cutting proceeds stepwise under safety gates (move $\rightarrow$ align $\rightarrow$ cut), policy computation constitutes only a small fraction of total episode time, indicating that current latency is unlikely to bottleneck clinical workflow. Nonetheless, further acceleration via distillation, low-bit quantization, and speculative decoding is under consideration.

\begin{table}[t]
\centering
\caption{Comparison of inference latency and robot execution time (mean) between the Vanilla model and the proposed ArthroCut framework.}
\label{tab:efficiency}
\setlength{\tabcolsep}{8pt}
\renewcommand{\arraystretch}{1.3}
\begin{tabular}{lcc}
\hline
Method           & Inference time (ms) & Robot exec. time (s) \\ \hline
Vanilla Model    & 267.14                 & 121.89                  \\
ArthroCut (Ours) & 302.52                 & 107.33                  \\ \hline
\end{tabular}
\end{table}

\begin{table}[t]
\centering
\caption{Ablation study on the proposed ArthroCut framework. PIT denotes the proposed preoperative imaging tokens and TAST denotes the time-aligned surgical tokens.}
\label{ab_study}
\renewcommand{\arraystretch}{1.2}
\resizebox{7cm}{!}{%
\begin{tabular}{lcccc}
\hline
\multirow{2}{*}{Method} & \multicolumn{2}{c}{Ablation}                         & \multirow{2}{*}{SR $\uparrow$} & \multirow{2}{*}{SPL $\uparrow$} \\
                        & PIT                       & TAST                      &                                &                                 \\ \hline
Vanilla Model                &                           &                           & 0.05                           & 0.04                            \\
Vanilla Model                 & \checkmark &                           & 0.12                           & 0.10                            \\
Vanilla Model                 &                           & \checkmark & 0.68                           & 0.61                            \\
ArthroCut (Ours)               & \checkmark & \checkmark & 0.86                           & 0.75                            \\ \hline
\end{tabular}
}
\end{table}

\section{Conclusion, Limitation and Future work}

In this work, we introduced \textit{ArthroCut}, an autonomous policy learning framework that translates time-aligned multimodal observations into structured and verifiable robot actions for multi-plane bone cutting in unicompartmental knee arthroplasty. The framework leverages two complementary token families: Preoperative Imaging Tokens (PIT), which compactly encode patient-specific CT/MR anatomy and surgical intent, and Time-Aligned Surgical Tokens (TAST), which fuse intraoperative RGB--D cues, SE(3) bone--tool relations from optical tracking, robot states, and the preceding action. Built on a Qwen2.5--VL--32B--Instruct backbone, the model performs unified next-token reasoning and generates an action grammar with quantized parameters under grammar and safety masks. Using a time-synchronized dataset of 21 cases (including 23{,}205 RGB--D pairs) and a physical bench evaluation with a knee prosthesis, \textit{ArthroCut} consistently outperforms state-of-the-art methods on \emph{Success Rate (SR)} and \emph{SPL}. Ablation studies further confirm that TAST is the primary driver of execution reliability, while PIT provides essential anatomical grounding; their integration yields the largest improvements and the most stable multi-plane performance. Furthermore, ArthroCut achieves a mean per-action latency of 302.52\,ms on a single A100 and reduces robot execution time by 11.95\% relative to the vanilla baseline.

Despite these encouraging results, several practical constraints remain. First, the dataset—though fully synchronized and multimodal—remains of moderate scale and is derived from a single collection pipeline, potentially limiting generalization to diverse anatomies, implant types, or surgical workflows. Second, evaluation was performed primarily on prosthesis-based bench tests; while providing controlled realism, these settings do not fully capture the complexity of live operating rooms (e.g., soft-tissue interference, blood, and intraoperative variability). Third, the current policy assumes calibrated RGB--D sensing and stable optical tracking, rendering it sensitive to sensor drift or temporary marker loss. These limitations are engineering in nature rather than fundamental, and they highlight clear avenues for enhancing robustness and external validity. Looking forward, we believe recent advances in autonomous surgical robots and generative models~\cite{kim2025srt, song2025rationalvla, long2025surgical} present promising opportunities to enhance surgical decision-making accuracy and surgical continuity.

\clearpage

\bibliographystyle{IEEEtran}
\bibliography{references}

\end{document}